\documentclass[sigconf]{acmart}

\usepackage{makecell}
\usepackage{float} 
\usepackage{bm}
\usepackage{graphicx}
\usepackage{booktabs}
\usepackage{multirow}
\usepackage{enumitem}
\setlist[itemize]{leftmargin=*}
\setlist[enumerate]{leftmargin=*}
\definecolor{lightgray}{RGB}{215,215,215}
\definecolor{lightblue}{RGB}{107,174,214}
\definecolor{bluu}{HTML}{ECF4FF}
\definecolor{blu}{RGB}{158,202,225}
\definecolor{myorange}{RGB}{2, 142, 2}
\usepackage{colortbl}  
\usepackage{color}
\usepackage{xcolor}
\usepackage[normalem]{ulem}
\useunder{\uline}{\ul}{}
\usepackage{subfigure}
\usepackage{wrapfig}
\usepackage{amsmath}

\usepackage[most]{tcolorbox}
\tcbuselibrary{listings}
\newtcblisting{promptbox}[1][]{
  colback=gray!5,       
  colframe=gray!40,     
  listing only,
  breakable,            
  pad at break*=2mm,    
  listing options={
    basicstyle=\ttfamily\footnotesize,
    columns=fullflexible,
    keepspaces=true,
    breaklines=true,
    language=,          
    xleftmargin=0pt,       
    breakindent=0pt,       
    showstringspaces=false,
    aboveskip=0pt,         
    escapeinside={*@}{@*}, 
  },
  left=6pt,
  right=6pt,
  top=6pt,
  bottom=6pt,
  title=#1,
  fonttitle=\bfseries,
}

\definecolor{myblue}{HTML}{228be6}
\definecolor{mygreen}{HTML}{40c057}
\definecolor{mygray}{HTML}{adb5bd}
\definecolor{darkgray}{HTML}{495057}

\tcbset{
    mydoublebox/.style={
        colback=white,     
        colframe=mygray,        
        fonttitle=\bfseries,    
        coltitle=white,         
        boxrule=0.5pt,          
        width=0.48\textwidth,       
        top=2mm, bottom=2mm,    
        left=2mm, right=2mm,   
        enhanced,                
        breakable,
        unbreakable,
    }
}

\newcommand{\sss}[1]{\subsubsection{\textbf{#1}}}

\newcommand{\ie}{\emph{i.e., }}
\newcommand{\eg}{\emph{e.g., }}

\copyrightyear{2026}
\acmYear{2026}
\setcopyright{cc}
\setcctype{by}
\acmConference[KDD '26]{Proceedings of the 32nd ACM SIGKDD Conference on Knowledge Discovery and Data Mining V.2}{August 09--13, 2026}{Jeju Island, Republic of Korea}
\acmBooktitle{Proceedings of the 32nd ACM SIGKDD Conference on Knowledge Discovery and Data Mining V.2 (KDD '26), August 09--13, 2026, Jeju Island, Republic of Korea}
\acmDOI{10.1145/3770855.3817545}
\acmISBN{979-8-4007-2259-2/2026/08}

\begin{document}

\title{Can Large Language Models Derive New Knowledge? A Dynamic Benchmark for Biological Knowledge Discovery}

\author{Chaoqun Yang}
\email{chaoqun@yang.email.cn}
\affiliation{
  \institution{National University of Singapore}
  \city{Singapore}
  \country{Singapore}
}

\author{Xinyu Lin}
\authornote{Corresponding authors.}
\email{xylin1028@gmail.com}
\affiliation{
  \institution{National University of Singapore}
  \city{Singapore}
  \country{Singapore}
  \country{}
}

\author{Shulin Li}
\authornotemark[1]
\email{lsl19@tsinghua.org.cn}
\affiliation{
  \institution{Tsinghua University}
  \city{Beijing}
  \country{China}
}

\author{Wenjie Wang}
\email{wenjiewang96@gmail.com}
\affiliation{
  \institution{University of Science and Technology of China}
  \city{Hefei}
  \country{China}
}

\author{Ruihan Guo}
\email{grhan@mail.ustc.edu.cn}
\affiliation{
  \institution{University of Science and Technology of China}
  \city{Hefei}
  \country{China}
}

\author{Fuli Feng}
\email{fulifeng93@gmail.com}
\affiliation{
  \institution{University of Science and Technology of China}
  \city{Hefei}
  \country{China}
}

\author{Tat-Seng Chua}
\email{dcscts@nus.edu.sg}
\affiliation{
  \institution{National University of Singapore}
  \city{Singapore}
  \country{Singapore}
}

\renewcommand{\shortauthors}{Chaoqun Yang et al.}

\begin{abstract}
Recent advancements in Large Language Model (LLM) agents have demonstrated remarkable potential in automatic knowledge discovery. However, rigorously evaluating an AI’s capacity for knowledge discovery remains a critical challenge. Existing benchmarks predominantly rely on static datasets, leading to inevitable data contamination where models have likely seen the evaluation knowledge during training. Furthermore, the rapid release cycles of modern LLMs render static benchmarks quickly outdated, failing to assess the ability to discover truly new knowledge. To address these limitations, we propose DBench-Bio, a dynamic and fully automated benchmark designed to evaluate AI’s biological knowledge discovery ability. DBench-Bio employs a three-stage pipeline: (1) data acquisition of rigorous, authoritative paper abstracts; (2) QA extraction utilizing LLMs to synthesize scientific hypothesis questions and corresponding discovery answers; and (3) QA filter to ensure quality based on relevance, clarity, and centrality. We instantiate this pipeline to construct a monthly-updated benchmark covering 12 biomedical sub-domains. Extensive evaluations of SOTA models reveal current limitations in discovering new knowledge. Our work provides the first dynamic, automatic framework for assessing the new knowledge discovery capabilities of AI systems, establishing a living, evolving resource for AI research community to catalyze the development of knowledge discovery. 

\end{abstract} 




\maketitle

\section{Introduction}

\begin{figure*}[t]
    \setlength{\belowcaptionskip}{0cm}
    \includegraphics[width=0.9\linewidth]{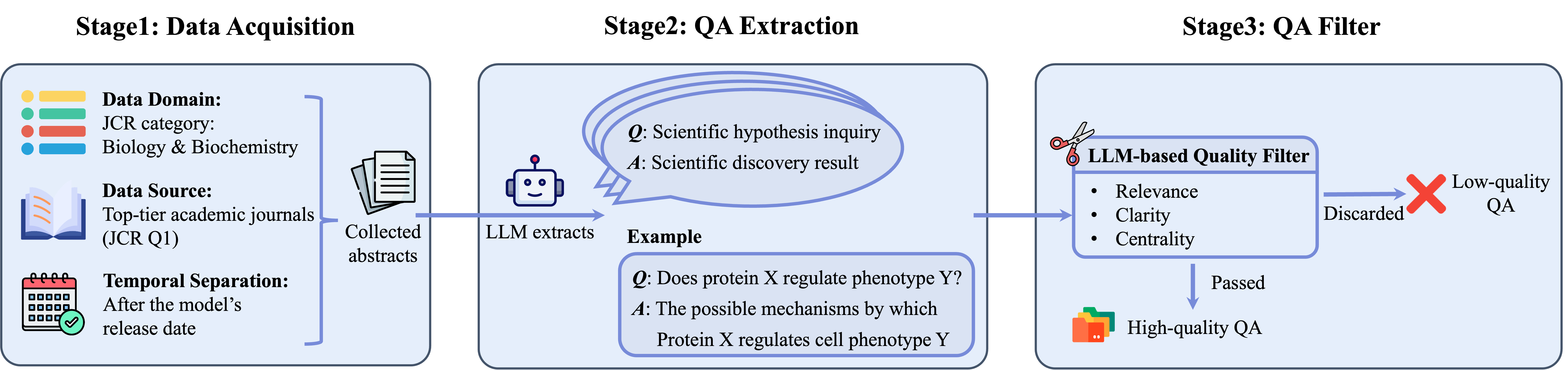}
    \caption{The overall pipeline of DBench-Bio, which consists of three stages. (1) Data Acquisition: We source abstracts from JCR Q1 ``Biology \& Biochemistry'' journals published post-model release to ensure rigor and prevent data leakage. (2) QA Extraction: Utilizing LLMs, we synthesize new knowledge into pairs consisting of scientific hypothesis questions and corresponding discovery answers. (3) QA Filter: We employ an LLM-based filter to remove low-quality pairs based on relevance, clarity, and centrality (see Section~\ref{sec:quality_assurance} for details).}
    \label{fig:method}
\end{figure*}

New knowledge discovery in the biological field serves as a cornerstone of modern healthcare, driving critical advancements in understanding disease mechanisms and developing therapies \cite{jumper2021highly, gao2024empowering}. However, the process of knowledge discovery is inherently labor-intensive and time-consuming, involving fragmented and complex workflows, such as literature review, hypothesis generation, experimental design, and data analysis. 
Recent advancements of Large Language Model (LLM) agents suggest a promising direction for partially automating such workflows.
Emerging systems, \eg AI co-scientist~\cite{gottweis2025towards}, Biomini~\cite{huang2025biomni}, and BioLab~\cite{jin2025biolab}, are capable of autonomously executing the entire knowledge discovery pipeline with minimal human intervention, signaling a new era of automated knowledge discovery.

As these AI systems advance, how to rigorously evaluate their capacity for new knowledge discovery becomes a critical problem. 
In response, many benchmarks have been proposed, such as Auto-Bench~\cite{chen2025auto}, ResearchBench~\cite{liu2025researchbench}, LLM-SRBench~\cite{shojaee2025llmsrbench}, and HiSciBench~\cite{zhang2025hiscibench}.
Although these benchmarks try to assess knowledge discovery ability, they are predominantly built upon static datasets collected at fixed time points.
This reliance introduces a fatal flaw: \textit{\textbf{data contamination}}, where the models may have already ``seen'' the knowledge contained in these existing benchmarks during training. 
Consequently, it becomes difficult to determine whether a model is genuinely discovering new knowledge or merely recalling previously seen information. This ambiguity severely hinders the advancement of AI for knowledge discovery.

To combat this issue, a fundamental prerequisite of knowledge discovery benchmark is: temporal separation, where the new knowledge in evaluation data must be temporally isolated from the model's training set, \ie the new knowledge should post-date the model's release time, to ensure the knowledge has not been seen by the LLMs\footnote{In case the search tools can be used by LLM agents, the searched items should also be restricted to those before the model release time.}. 
Nonetheless, enforcing temporal separation in practice is challenging because modern LLMs are continuously retrained on new web data over time and frequently release new versions (\eg GPT-5.2 had been released merely one month after GPT-5.1~\cite{openai2025introducing}).
Although static benchmarks can enforce temporal separation at creation, they quickly become outdated as models evolve. 
Therefore, to reliably assess knowledge discovery, benchmarks must be \textbf{dynamic}, continuously evolving to guarantee the knowledge remains new for released LLMs. 

However, constructing high-quality benchmarks traditionally relies on manual curation—a process including data collection, QA design, and quality inspection, which is intrinsically \textit{labor-intensive} and costly. 
It is virtually impossible in practice to sustain rapid benchmark iterations through manual effort. 
To pursue rapid iteration, we need an efficient benchmark construction methodology. 
A promising approach is to leverage LLMs to extract knowledge discovery QA pairs from newly published research articles, given their strong comprehension and generation capabilities~\cite{long2024llms}. 
Nonetheless, this strategy introduces potential quality risks that must be carefully addressed both before and after QA extraction: 
\begin{itemize}
    \item Pre-extraction: Scientific literature is vast and heterogeneous, and may contain a substantial proportion of low-quality or irrelevant publications. 
    Thus, it is crucial to acquire scientifically rigorous, highly authoritative, and temporally isolated data sources, allowing LLMs to generate high-quality knowledge QA pairs for reliable evaluation. 
    \item Post-extraction: LLMs are prone to instruction-following failures that can compromise data quality. For example, the focus of the generated question may not match the expected topic, or the generated questions may emphasize peripheral experimental details rather than core scientific findings. 
    As such, it is necessary to design a rigorous quality assurance mechanism to filter out these low-quality extracted data. 
\end{itemize}

To this end, we propose \textbf{DBench-Bio}, a \textbf{D}ynamic and fully automated \textbf{Bench}mark designed to evaluate AI's \textbf{Bio}logical knowledge discovery ability. 
DBench-Bio focuses on derivation-based knowledge discovery, defined as the ability to combine existing, well-established knowledge to logically deduce and infer new knowledge. This formulation is motivated by the process of hypothesis generation in scientific discovery: forming a novel scientific hypothesis is essentially a reasoning process that bridges and recombines prior knowledge to explain unexplored phenomena \cite{hanson1979patterns, dunbar1995scientists}.

As shown in Figure~\ref{fig:method}, DBench-Bio employs a three-stage pipeline to construct the benchmark: (1) Data Acquisition: For pre-extraction, to ensure the scientific rigor, we source publications from the ``Biology \& Biochemistry'' category of the Journal Citation Reports (JCR) system. To uphold authority, we exclusively include articles from journals ranked in the JCR Q1 tier. Furthermore, to guarantee strict temporal separation and prevent data contamination, we crawl abstracts of target journals published only \textit{after} the model's release date, thereby capturing the latest human knowledge discoveries. 
(2) QA Extraction: To enable efficient construction, we leverage LLMs to synthesize new knowledge discovery QA pairs, where questions represent scientific hypothesis inquiries (\eg ``Does protein X regulate cell phenotype Y?''), and answers represent the corresponding discovery results (\eg ``The possible mechanisms by which Protein X regulates cell phenotype Y.''). 
(3) QA Filter: For post-extraction, we employ an LLM-based filter to eliminate low-quality pairs based on three quality indicators: relevance, clarity, and centrality (see Section~\ref{sec:quality_assurance} for details). 

We instantiate this pipeline to construct a monthly-updated benchmark and conduct extensive evaluations of SOTA models, including three variants (\ie SOTA LLMs with and without thinking, with tool-using, and with agentic workflow).\footnote{Code is available at \url{https://github.com/transcend-0/DBench-Bio}.}
Our results reveal the current limitations of LLMs in discovering new knowledge:
(1) While existing models excel at retrieval, their capacity for genuine discovery remains weak, particularly within the domain of mathematical biology. (2) Even when models demonstrate a strong grasp of fundamental biological concepts, they perform poorly in discovering new biological knowledge. This implies that knowledge discovery is a capability that cannot be sufficiently acquired simply by scaling up pre-training data. Instead, it necessitates specialized mechanisms for further development. (3) Through a systematic analysis of failure cases, we identify five primary modes of error: mechanism errors, information omission, generic mechanism substitution, refusal to answer, and overconfident reasoning.


Our contributions are summarized as follows:
\begin{itemize}
    \item We introduce the first methodology for constructing a dynamic, fully automatic benchmark for assessing AI's new knowledge discovery ability.
    \item We instantiate this methodology to create and release a monthly-updated benchmark in the biological sciences, including 12 sub-domains, providing a living resource for the community.
    \item We conduct extensive empirical evaluations using our benchmark, offering quantitative insights into the capabilities and limitations of current SOTA AI models when confronted with cutting-edge, discovery-oriented questions.
\end{itemize}

\section{Related Work}

\subsection{Scientific Benchmark}

Scientific benchmarks, aimed at rigorously assessing AI capabilities in scientific understanding and application, have evolved significantly over the years. 
Early efforts primarily focused on evaluating domain knowledge and foundational understanding by extracting problems from educational resources. Benchmarks such as SuperGLUE~\cite{wang2019superglue}, MMLU~\cite{hendrycks2021mmlu}, MMLU-Pro~\cite{wang2024mmlu}, MMMU~\cite{yue2024mmmu}, C-Eval~\cite{huang2023c}, Xiezhi~\cite{gu2024xiezhi}, GPQA~\cite{rein2024gpqa}, and HLE~\cite{phan2025humanity} typically construct multiple-choice or short-answer questions derived from textbooks, standardized exams, or academic papers. While these datasets effectively measure a model’s ability to retain and apply established scientific facts, they are limited in assessing complex problem-solving capabilities.
To bridge the gap between simple knowledge understanding and complex reasoning, recent efforts have sought to increase the difficulty and cognitive load of evaluations. For instance, SciEval~\cite{sun2024scieval} assesses AI's problem-solving capabilities across various disciplines from four distinct dimensions: basic knowledge, knowledge application, scientific calculation, and research ability. Concurrently, OlympicArena~\cite{huang2024olympicarena} aggregates challenging problems across various scientific disciplines from Olympic competitions. However, these benchmarks, while challenging, primarily evaluate the LLM's mastery of known knowledge.
More recently, the focus has shifted towards evaluating the LLM's ability to explore unknown knowledge, that is, knowledge discovery capability. Current benchmarks in this domain target specific aspects of the discovery process: Auto-Bench~\cite{chen2025auto} evaluates a model’s ability to discover knowledge within causal graphs; ResearchBench~\cite{liu2025researchbench} assesses the capacity to propose scientific hypotheses; LLM-SRBench~\cite{shojaee2025llmsrbench} focuses on scientific equation discovery; and HiSciBench~\cite{zhang2025hiscibench} evaluates knowledge discovery through data analysis.

Despite these advancements in assessing knowledge discovery, existing methods rely heavily on static datasets. This reliance introduces significant issues regarding data contamination, as LLMs may have simply memorized the answers during training, thereby ``cheating'' to achieve artificially high scores. In contrast to these static approaches, our proposed DBench-Bio is designed not only to evaluate scientific reasoning but also to ensure the integrity of the assessment by mitigating contamination through dynamic updates.

\subsection{Dynamic Benchmark}

Traditional benchmarks for evaluating AI systems are typically static: a fixed dataset is released once and reused indefinitely~\cite{chen2025benchmarking}. This approach suffers from a critical limitation: data contamination, where test samples leak into training corpora, leading to inflated performance. In response, recent research has proposed ``dynamic'' benchmarks, which continuously introduce fresh or synthetically modified instances that are guaranteed to be absent from training data~\cite{chen2025benchmarking}. 
Benchmarks such as LiveBench~\cite{white2025livebench}, LiveCodeBench~\cite{jain2025livecodebench}, ForecastBench~\cite{karger2025forecastbench}, and FutureX~\cite{zeng2025futurex} automatically crawl the latest web data, dynamically build questions, and update monthly or weekly to enable more reliable, contamination-free evaluations.

However, existing dynamic benchmarks primarily focus on domains such as code programming,  event forecasting, and general reasoning, and there has not yet been a dynamic benchmark specifically tailored to evaluate knowledge discovery. 
Our DBench-Bio addresses this gap by introducing the first dynamic framework designed to assess AI's capabilities in knowledge discovery.

\subsection{Automated Benchmark Construction}

Traditional benchmarks rely heavily on manual curation, which ensures quality but is costly, hard to scale, and prone to saturation as models improve. In response, several lines of work have introduced automated pipelines that leverage templates, LLMs, and web agents to construct benchmarks. 
GSM-Symbolic~\cite{mirzadeh2024gsm} and Mathador-LM~\cite{kurtic2024mathador} use question templates with placeholder variables to randomly generate math problems. SciQAG~\cite{wan2024sciqag} leverages LLMs to extract diverse and research-level questions and answers from scientific papers.
TAM-Bench~\cite{jia2025towards} employs web agents to automatically collect and structures machine learning challenges from platforms such as Kaggle.

Collectively, these efforts demonstrate the efficacy of automated pipelines in generating diverse and scalable evaluation data. Unlike existing benchmarks that primarily focus on generating diverse test items, knowledge discovery demands strict temporal isolation and rigorous quality control. However, at present, there is no automated benchmark specifically designed for knowledge discovery, which is exactly what our proposed DBench-Bio seeks to contribute. 

\section{DBench-Bio}

To construct a rigorous benchmark for evaluating knowledge discovery capabilities in the biology domain, we adopt a systematic pipeline comprising three distinct stages: data acquisition, QA extraction, and QA filter (see Figure~\ref{fig:method}). This pipeline is designed to ensure the authority, recency, and high quality of our benchmark.

\subsection{Data Acquisition}

To ensure the quality of the data, we adopt three selection criteria: data domain, data source, and temporal separation.
(1) For data domain, to achieve comprehensive yet representative coverage of the domain, we select based on the Journal Citation Reports (JCR) domain classification system\footnote{\url{https://jcr.clarivate.com/jcr/browse-categories}.}. Using the JCR taxonomy provides a methodological advantage by relying on a standardized, widely accepted framework for categorizing scientific impact. Specifically, we identify the ``Biology \& Biochemistry'' category within the JCR, from which a domain expert carefully selects 12 representative sub-domains related to cell biology. This selection is intended to capture the diversity of the field while maintaining a manageable scope, and excludes less relevant areas (\eg Agronomy, Zoology).
(2) For data source, to further ensure authority, we adapt an impact-based filter. Within each selected sub-domain, we restrict our collection to journals ranked in the Journal Impact Factor (JIF) Quartile Q1. 
(3) To guarantee strict temporal separation and prevent data contamination, we crawl abstracts of target journals published only after the model's release date\footnote{To minimize data contamination risks, we deliberately use PubMed's ``Entry Date'' rather than the ``Publication Date'', as the former better approximates the moment a paper enters the public domain.}. 
We focus on abstracts rather than full texts, as abstracts provide dense, high-level summaries of novel insights and scientific findings, minimizing noise from extraneous experimental details.

By employing these data acquisition strategies, we construct a benchmark that supports monthly updates. As the initial implementation of this pipeline, we focus our data collection on research publications from December 2025 and January 2026.

\subsection{QA Extraction}

To transform these unstructured paper abstracts into a structured benchmark format, we employ a capable LLM (\eg DeepSeek-V3.2-thinking) to extract QA pairs. Each abstract corresponds to one QA pair. We design specific prompts to instruct the model: the \textit{Question} must represent a precise scientific inquiry, while the corresponding \textit{Answer} must encapsulate the core conclusion of the scientific finding derived directly from the abstract. 

Crucially, these QA pairs are designed to focus on knowledge discovery and high-level concepts, avoiding fine-grained details such as specific parameters or numerical values. Specifically, we prioritize causal inference (\eg ``Does X regulate Y phenomenon?'' or ``What are the implications of Z on disease progression?'') and mechanistic explanation (\eg ``How does X regulate Y phenomenon?'' or ``What is the mechanism behind Z effect?''). Furthermore, we ensure that the language is fluent and natural, with answers formatted as clear and concise bullet points.
Full prompts can be found in Appendix~\ref{app:prompts}.

\begin{table}[t]
    \setlength{\abovecaptionskip}{0cm}
    \setlength{\belowcaptionskip}{0cm}
    \centering
    \caption{The Alt-test~\cite{calderon-etal-2025-alternative} results and inter-annotator agreement results. The winning rates for all metrics surpass 0.5, achieving statistical significance. Furthermore, the advantage probabilities consistently surpass 0.8, suggesting a high likelihood that LLM evaluators can serve as viable substitutes for human expert assessors. We also report the inter-annotator agreement using Gwet’s AC1, with all scores exceeding 0.6, demonstrating substantial agreement among experts.}
    \setlength{\tabcolsep}{1.3mm}
    \begin{tabular}{r|c|c|c|c}
    \toprule
    ~ & \textbf{Relevance} & \textbf{Clarity} & \textbf{Centrality} & \textbf{Evaluation} \\ \hline
    \makecell[c]{\textbf{Winning} \\ \textbf{Rate}} & 0.90 & 1.00 & 1.00 & 0.90 \\ \hline
    \makecell[c]{\textbf{Advantage} \\ \textbf{Probability}} & 0.94 & 0.96 & 1.00 & 0.82 \\ \hline
    \makecell[c]{\textbf{Gwet's} \\ \textbf{AC1}} & 0.98 & 0.93 & 0.99 & 0.78 \\
    \bottomrule
    \end{tabular}
    \label{tab:human_test}
\end{table}

\subsection{QA Filter}
\label{sec:quality_assurance}

Acknowledging that LLMs may occasionally exhibit instruction-following failures or hallucinations, we implement a robust, multi-dimensional quality control mechanism to ensure the benchmark's reliability. Inspired by previous work~\cite{gu2024survey,li2024llms}, we utilize an LLM as a judge to assess each generated QA pair.

Each QA pair is assessed across three distinct dimensions, scored on a discrete scale from 1 to 5:

\begin{itemize}
    \item \textbf{Relevance:} This metric measures the semantic alignment between the QA pair and the target biology sub-domain. Even within Q1 journals, article scopes may diverge from the specific taxonomy used for benchmark categorization. This step filters out such misaligned content.
    \item \textbf{Clarity:} This dimension evaluates the linguistic precision and coherence of both the question and the answer, ensuring that the question is unambiguous and the answer is articulated clearly. Notably, it screens for context-dependent phrases (e.g., ``based on the text'') that would effectively degenerate the task into a reading comprehension exercise rather than an independent knowledge assessment.
    \item \textbf{Centrality:} To prevent the benchmark from focusing on trivial details, this metric assesses whether the QA pair corresponds to the primary scientific finding of the abstract, rather than peripheral or background information.
\end{itemize}

To filter low-quality data in the benchmark, we enforce a strict quality threshold. Only QA pairs that meet the following criteria are retained in the final dataset: Relevance $\ge$ 4, Clarity $\ge$ 5, and Centrality $\ge$ 5. This rigorous standard ensures that the benchmark consists exclusively of high-quality, core scientific knowledge.

\textbf{Consistency with Human Experts.} Following the guidelines of Alt-test\cite{calderon-etal-2025-alternative}, we engage three human experts who possess in-depth professional knowledge in biology to manually assess 200 QA pairs, aiming to quantify the alignment between LLMs and human judgment. As shown in Table~\ref{tab:human_test}, the winning rates all surpass 0.5, passing the statistical test. Additionally, with advantage probabilities exceeding 0.8, our results provide strong evidence that LLM annotation serves as a viable substitute for human annotation. We also report the inter-annotator agreement using Gwet’s AC1, with all scores exceeding 0.6, demonstrating substantial agreement among experts.

\subsection{Evaluation Protocol}

Following the previous work~\cite{gu2024survey,li2024llms}, we employ an ``LLM-as-a-Judge'' framework to evaluate the performance of target models. The primary objective is to assess the consistency between the answers generated by the models and the ground-truth answers in our benchmark.

A judge LLM is provided with the gold standard answer and the model's generated answer. The judge assigns a score ranging from 1 to 5 on a Likert scale, where 5 indicates a correct and comprehensive answer that fully matches the gold standard, while 1 represents a completely incorrect, hallucinated, or irrelevant response.

To ensure the credibility of the LLM assessment, we do the Alt-test~\cite{calderon-etal-2025-alternative}. Three human experts are engaged to manually evaluate the consistency between the correct answers and the reference answers. As shown in Table~\ref{tab:human_test}, the winning rate surpasses 0.5, passing the statistical test. Additionally, with an advantage probability exceeding 0.8, our results provide strong evidence that LLM annotation serves as a viable substitute for human evaluator.

\begin{figure*}[t]
    \setlength{\abovecaptionskip}{0cm}
    \setlength{\belowcaptionskip}{0cm}
    \includegraphics[width=1\linewidth]{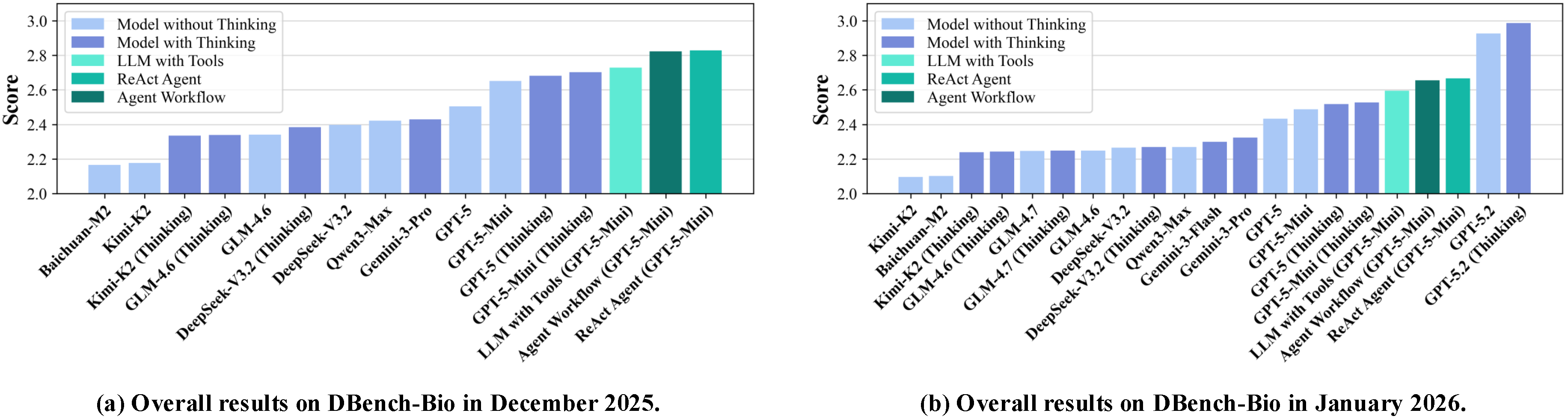}
    \caption{Overall results on DBench-Bio.}
    \label{fig:result}
\end{figure*}

\begin{figure}[t]
    \setlength{\abovecaptionskip}{0cm}
    \setlength{\belowcaptionskip}{0cm}
    \includegraphics[width=1\linewidth]{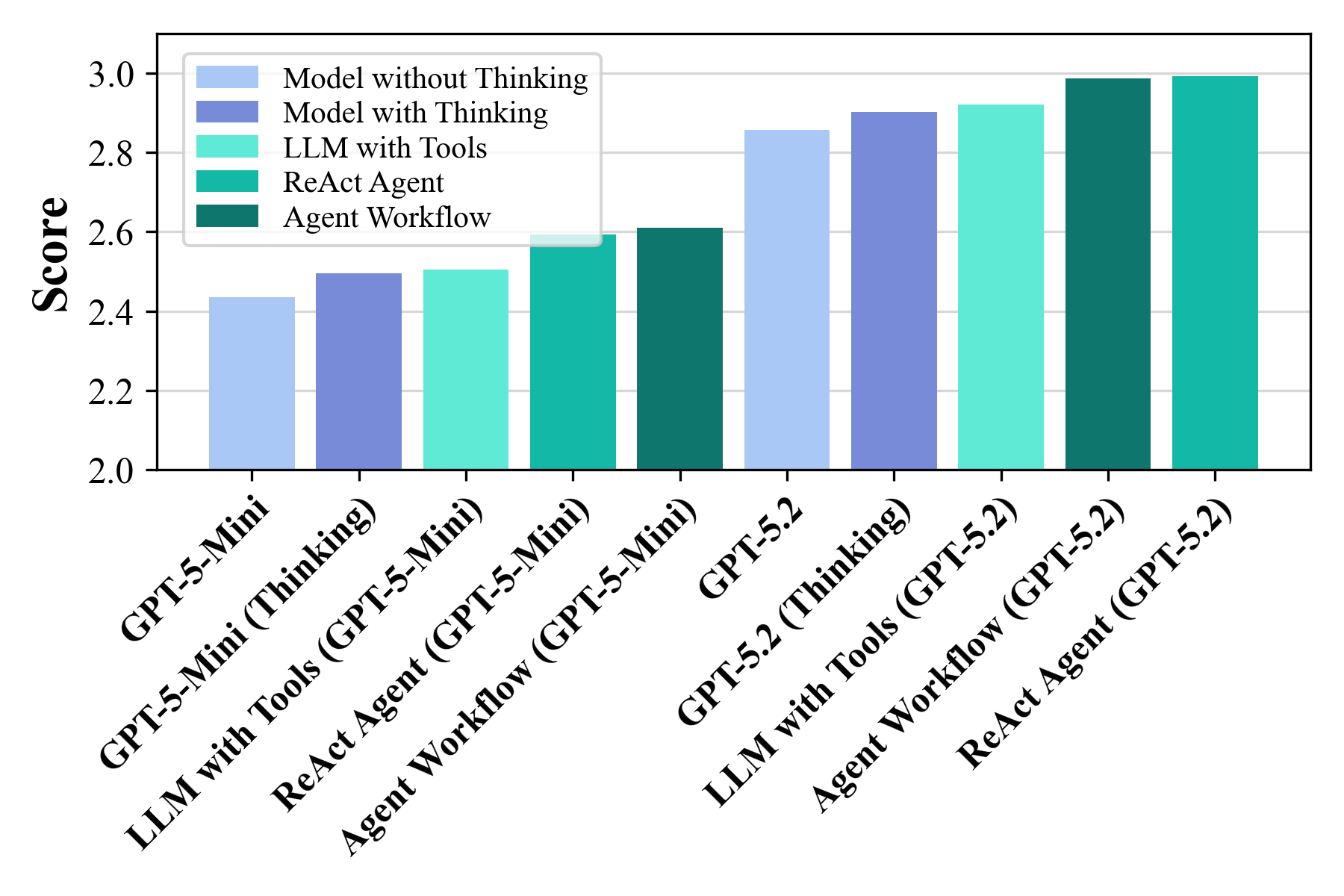}
    \caption{Results for agent-based methods on DBench-Bio.}
    \label{fig:result_gpt}
\end{figure}

\begin{figure*}[t]
    \setlength{\abovecaptionskip}{0cm}
    \setlength{\belowcaptionskip}{0cm}
    \includegraphics[width=1\linewidth]{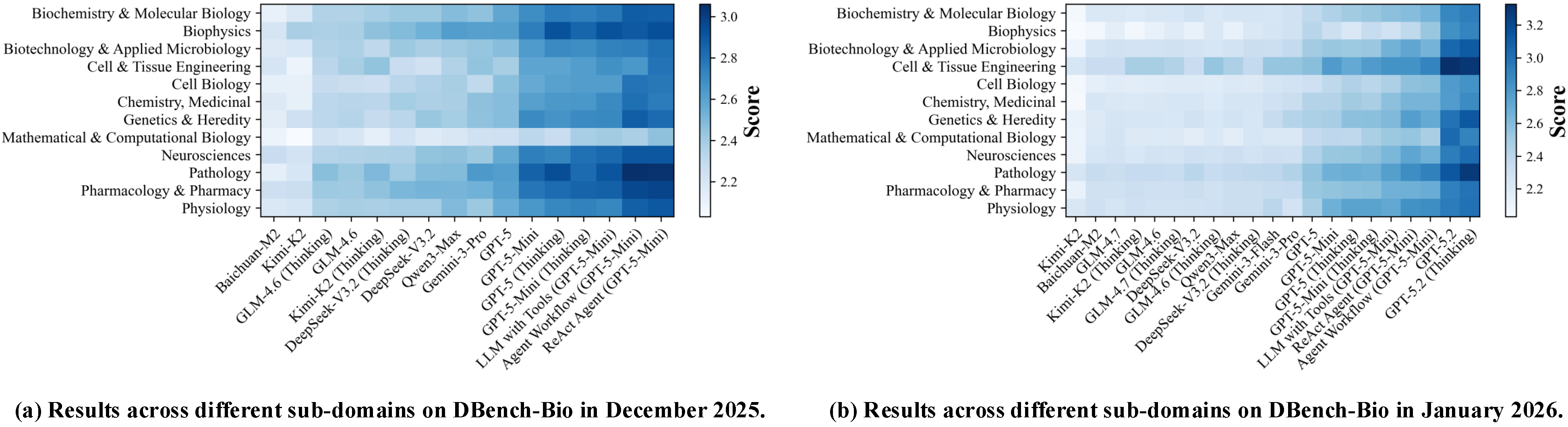}
    \caption{Results across different domains on DBench-Bio.}
    \label{fig:category}
\end{figure*}

\begin{figure}[t]
    \setlength{\abovecaptionskip}{0cm}
    \setlength{\belowcaptionskip}{0cm}
    \includegraphics[width=1\linewidth]{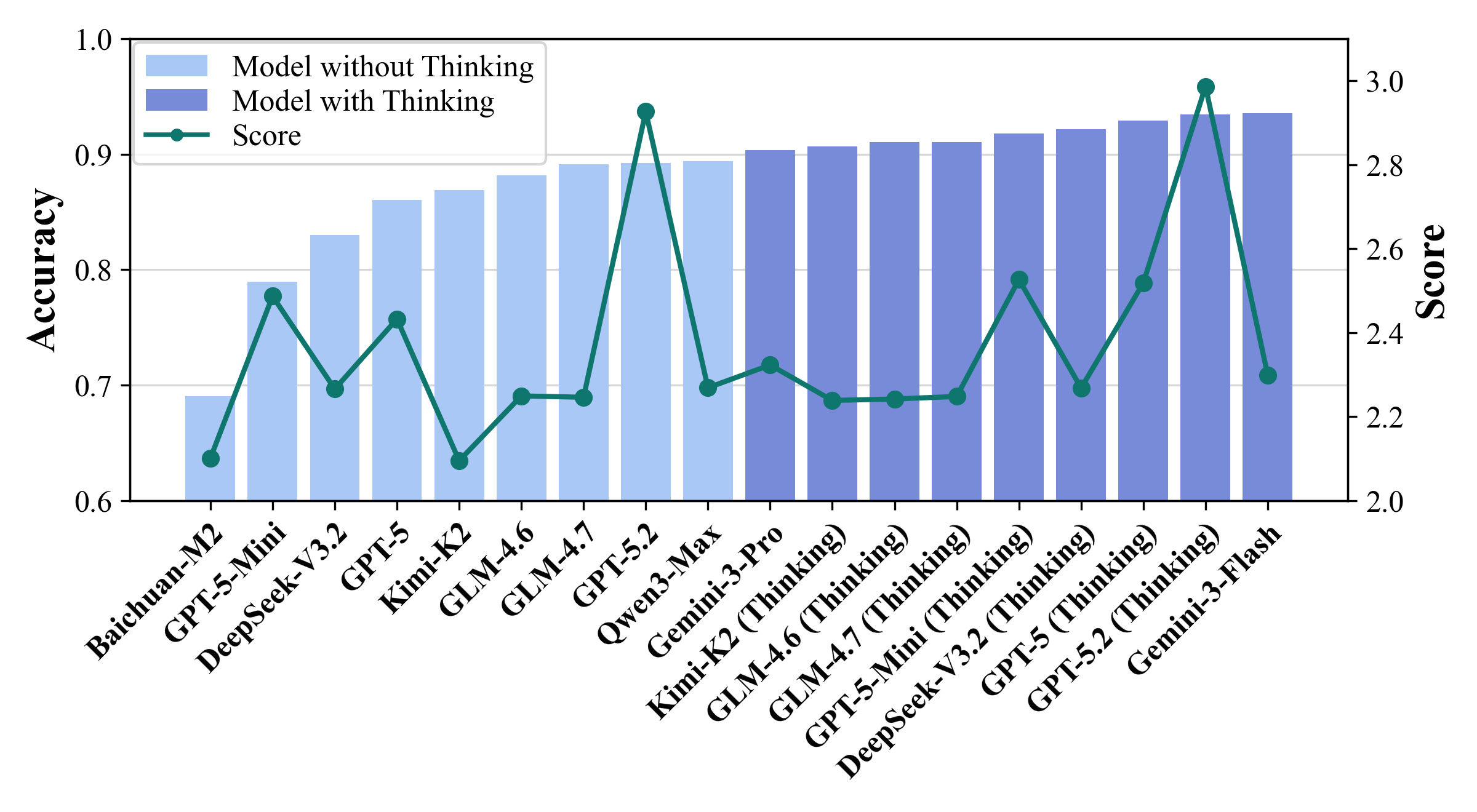}
    \caption{Results for base models on MMLU-Pro (Biology) (bar chart) and DBench-Bio (January 2026) (line graph).}
    \label{fig:MMLU-Pro}
\end{figure}

\section{Experiments}

In this section, we conduct extensive experiments to answer the following research questions: 
\begin{itemize}
    \item \textbf{RQ1}: Can SOTA LLMs and LLM Agents derive new biological knowledge? 
    \item \textbf{RQ2}: How do different models perform on different sub-domains?
    \item \textbf{RQ3}: How much does basic biological knowledge contribute to the discovery of new biological knowledge?
    \item \textbf{RQ4}: Why do LLMs fail to derive new knowledge?
\end{itemize}

\subsection{Experimental Setup}

\sss{Evaluated Models}
We evaluate the following methods on our proposed benchmark:
\begin{itemize}
    \item \textbf{Base LLMs:} A comprehensive selection of open-source and closed-source LLMs, encompassing both reasoning and non-reasoning models. These including GPT-5.2, GPT-5, GPT-5-Mini\footnote{GPT series model's non-thinking mode sets \textit{reasoning\_effort} to \textit{none} or \textit{minimal}.}, Gemini-3-Pro, Gemini-3-Flash, DeepSeek-V3.2, Kimi-K2, GLM-4.7, GLM-4.6, Qwen3-Max, and Baichuan-M2.
    \item \textbf{LLM with Tools:} A simple retrieval-augmented generation (RAG) system~\cite{arslan2024survey}, namely a base LLM augmented with a single tool-use capability.
    \item \textbf{ReAct Agent:} We employ the ReAct (Reasoning + Acting) paradigm \cite{yao2022react}, which operates on an iterative ``Reason-Act'' loop. In this framework, the model generates reasoning traces to plan actions, executes tool calls, and subsequently observes the results to inform the next step of reasoning until the task is resolved.
    \item \textbf{Agent Workflow:} A workflow-orchestrated multi-agent framework comprising specialized roles~\cite{jin2025biolab}: the \textit{Planner} analyzes input queries, decomposes them into sub-problems, and plans actions; the \textit{Tool Caller} invokes the relevant tools to collect information; the \textit{Reasoner} synthesizes the collected information; the \textit{Reporter} that formulates the final answer; and the \textit{Critic} that reviews the quality of the answer.
\end{itemize}

\sss{Implementation Details}

For methods that involve external tool integration (specifically \textit{LLM with Tools}, \textit{ReAct Agent}, and \textit{Agent Workflow}), we utilize GPT-5-Mini as the backbone model. The external tool provided is PubMed literature search. To prevent data leakage and ensure a rigorous evaluation, we strictly constrain the literature retrieval scope to publications released before the data collection phase of the benchmark. Specifically, for the December 2025 benchmark, the tool is restricted to retrieving literature dated before November 30, 2025.
Additionally, we use DeepSeek-V3.2 for both QA filtering and answer evaluation.
Unless otherwise specified, all models are evaluated using their default inference parameters. 

\subsection{Overall Results (RQ1)}

\begin{table}[t]
    \setlength{\abovecaptionskip}{0cm}
    \setlength{\belowcaptionskip}{0cm}
    \centering
    \caption{The data sample size and distribution.}
    \begin{tabular}{c|c|c}
    \toprule
    \textbf{Subdomain} & \textbf{202512} & \textbf{202601} \\ \hline
    \makecell[c]{\textit{Total}} & \textit{4231} & \textit{3791} \\ \hline
    \makecell[c]{Biochemistry \& Molecular Biology} & 500 & 500 \\
    \makecell[c]{Biophysics} & 79 & 32 \\
    \makecell[c]{Biotechnology \& Applied Microbiology} & 500 & 500 \\
    \makecell[c]{Cell \& Tissue Engineering} & 34 & 32 \\
    \makecell[c]{Cell Biology} & 500 & 500 \\
    \makecell[c]{Chemistry, Medicinal} & 500 & 500 \\
    \makecell[c]{Genetics \& Heredity} & 351 & 250 \\
    \makecell[c]{Mathematical \& Computational Biology} & 327 & 215 \\
    \makecell[c]{Neurosciences} & 500 & 500 \\
    \makecell[c]{Pathology} & 172 & 148 \\
    \makecell[c]{Pharmacology \& Pharmacy} & 500 & 500 \\
    \makecell[c]{Physiology} & 268 & 104 \\
    \bottomrule
    \end{tabular}
    \label{tab:data_size}
\end{table}

To validate the robustness of our methodology, we construct instances of DBench-Bio spanning two distinct temporal snapshots: December 2025 and January 2026.
The data sample size and distribution are shown in Table~\ref{tab:data_size}. Due to the inherent imbalance in the monthly publication volume across different subdomains, some fields naturally yield fewer QA pairs, while others yield an abundance. To ensure a balanced evaluation and prevent smaller domains from skewing the overall results, we apply a random sampling cap of 500 QA pairs for any subdomain that exceeded this threshold.
The overall results, illustrated in Figure~\ref{fig:result}, demonstrate consistent trends across both periods, leading to the following key observations:

\begin{itemize}
    \item \textbf{Low Overall Performance.} The aggregate performance across all evaluated models remains low (The maximum Score is 5). This underscores the inherent difficulty of knowledge discovery, suggesting that current LLM models have yet to master the ability to derive new knowledge.
    \item \textbf{Divergent Efficacy of Thinking Strategies.} Thinking strategies yield significant improvements for certain models (\eg Kimi-K2, GPT-5, GPT-5-Mini, and GPT-5.2), but offer negligible gains for others (\eg GLM-4.6, GLM-4.7, and DeepSeek-V3.2). This disparity indicates that different models possess varying reasoning capacities when tackling new knowledge problems. Furthermore, it implies that the utility of explicit thinking strategies is not universal but depends heavily on the model's intrinsic alignment with structured reasoning patterns.
    \item \textbf{Limited Utility of Tool Use.} Enabling tool use within a restricted retrieval scope failed to yield significant improvements. We attribute this to the fact that the information retrieved via tools largely overlaps with the models’ internal knowledge, thereby offering negligible information gain.
    \item \textbf{Effectiveness of Agent Architectures.} Both ReAct and Workflow architectures result in performance boosts. Interestingly, the performance gap between these two paradigms is marginal. This finding suggests that integrating reasoning with external tools, via either iterative planning or pre-set workflows, effectively facilitates the discovery of new knowledge.
\end{itemize}

In Figure~\ref{fig:result}, GPT-5.2 demonstrates superior performance over the agent-based methods. This discrepancy is likely attributed to the fact that our agent-based methods utilize GPT-5-Mini as the backbone model, which is significantly less capable than GPT-5.2. To verify this hypothesis, it is essential to investigate the impact of the backbone model on the performance of agent-based methods. However, given the high costs of GPT-5.2, we randomly sampled a subset of 1,000 instances for this experiment. The results are shown in Figure~\ref{fig:result_gpt}, from which we can observe that:
\begin{itemize}
    \item \textbf{Consistent Superiority of Agent Architectures.} Agent-based methods consistently outperform their corresponding backbone models. Notably, ReAct and Workflow achieve comparable performance, with negligible differences between them.
    \item \textbf{Scalability with Model Capacity.} The performance of agent-based methods scales positively with the capabilities of the backbone model. This observation indicates that the agent framework acts as an effective amplifier of the LLM's intrinsic potential. Stronger backbone models lead to more capable agents.
\end{itemize}

\subsection{Results Across Different Sub-domains (RQ2)}

Figure~\ref{fig:category} presents a domain-wise breakdown of model performance on DBench-Bio (December 2025 and January 2026), highlighting several critical trends:

\begin{itemize}
    \item \textbf{Cross-Domain Consistency.} The heatmap reveals a stable performance ranking across various sub-domains, with the GPT-5 series maintaining a consistent lead over other models. GPT-5 series consistently outperforms others across nearly every sub-domain, demonstrating a comprehensive capability that is insensitive to specific category shifts.
    \item \textbf{The Challenge of Mathematical Domains.} The ``Mathematical \& Computational Biology'' category emerges as the most intractable domain. Performance metrics for all models noticeably decline in this area, suggesting that current LLMs still struggle with tasks requiring complex mathematical reasoning and computational simulation.
\end{itemize}

\subsection{In-depth Analysis (RQ3)}

To investigate the contribution of basic biological knowledge to the discovery of new biological knowledge, we evaluate the base models on the biology task of MMLU-Pro benchmark~\cite{wang2024mmlu}, which offers a rigorous assessment of foundational biological literacy and textbook-level reasoning capabilities. We then compare these results with the evaluation results on DBench-Bio, as shown in Figure~\ref{fig:MMLU-Pro}. We can conclude the following observations:
\begin{itemize}
    \item \textbf{Basic Knowledge as a Prerequisite.} The mastery of basic knowledge serves as a prerequisite for discovering new knowledge. Notably, Baichuan-M2 exhibits the poorest performance on both MMLU-Pro and DBench-Bio, suggesting that deficiencies in basic knowledge retention severely hinder the capacity for new discovery.
    \item \textbf{The Gap Between Memorization and Discovery.} Thinking models consistently achieve accuracy exceeding 90\% on MMLU-Pro with negligible variance, indicating a robust grasp of biological basics. However, their performance on DBench-Bio remains suboptimal. This discrepancy suggests that high scores on MMLU-Pro may be artificially inflated by potential data contamination. More fundamentally, it highlights the limitations of current models in knowledge composition and complex reasoning. These findings imply that genuine knowledge discovery necessitates advanced reasoning capabilities, positioning DBench-Bio as a benchmark with superior discriminative power compared to traditional static alternatives.
    \item \textbf{Misalignment in Model Rankings.} We observe a misalignment in model rankings between MMLU-Pro and DBench-Bio. Despite the leading performance of Gemini-3-Flash on MMLU-Pro, its relatively inferior result on DBench-Bio suggests that high proficiency in static knowledge retention does not necessarily translate to the ability to discover new knowledge.
\end{itemize}

\subsection{Case Study (RQ4)}

By analyzing the failure cases of LLM, we identify that the main reasons for LLM's failure in inferring new knowledge fall into the following categories.

\textbf{Mechanism Error.} Although LLMs try their best to propose various possible mechanisms, none of these mechanisms is the correct answer. A typical case is as follows:
\begin{promptbox}[Question]
How does Brusatol (BRU) inhibit gastric cancer progression?
\end{promptbox}

\begin{promptbox}[Correct Answer]
1. BRU directly binds to and promotes the degradation of P4HA2, a key metabolic enzyme. 
2. This suppresses glycolysis, leading to reduced lactate production. 
3. Lower lactate levels inhibit histone lactylation modifications, specifically H3K9lac and H3K18lac, by affecting the lactylation writer HBO1.
4. The decline in histone lactylation represses the transcription of TTK, a mitotic regulator essential for cell division.
5. Ultimately, this disrupts the metabolic-epigenetic signaling cascade, inhibiting cancer cell proliferation and tumor growth.
\end{promptbox}

\begin{promptbox}[Model Response (Mechanism Error)]
1. Brusatol likely inhibits gastric cancer progression by suppressing the NRF2 antioxidant pathway, increasing oxidative stress and promoting apoptosis in cancer cells.
2. It may downregulate key oncogenic signaling pathways such as PI3K/AKT and MAPK, reducing cell proliferation and survival.
3. Brusatol could induce cell cycle arrest by modulating cyclin-dependent kinases and checkpoint proteins, halting tumor growth.
4. The compound might inhibit epithelial-mesenchymal transition by downregulating transcription factors like SNAIL and TWIST, reducing metastasis.
5. Brusatol potentially enhances chemosensitivity by inhibiting drug efflux pumps and DNA repair mechanisms, making cancer cells more vulnerable to treatment.
\end{promptbox}

\textbf{Information Omission.} The model identifies a partially correct mechanism or relevant biological process but omits key information required by the question. As a result, the answer remains incomplete.
A typical case is as follows:
\begin{promptbox}[Question]
What are the key cancer cell-autonomous mechanisms that contribute to immune evasion in cancer?
\end{promptbox}

\begin{promptbox}[Correct Answer]
1. Desensitization to cytokine signaling, such as interferon-gamma and tumor necrosis factor.
2. Impaired antigen presentation, reducing the visibility of cancer cells to the immune system.
3. Upregulation of suppressive ligands like programmed cell death ligand 1 (PD-L1) to inhibit T cell activity.
4. Epigenetic silencing of immunogenic pathways, limiting immune recognition and response.
\end{promptbox}

\begin{promptbox}[Model Response (Information Omission)]
1. Downregulation of MHC class I molecules impairs antigen presentation to CD8+ T cells.
2. Expression of immune checkpoint ligands like PD-L1 inhibits T cell activation and promotes exhaustion.
3. Defects in antigen processing machinery reduce tumor antigen visibility.
4. Secretion of immunosuppressive cytokines such as TGF-beta directly suppresses effector immune cells.
5. Enhanced expression of anti-apoptotic proteins confers resistance to immune-mediated cytotoxicity.
\end{promptbox}

\textbf{Generic Mechanism Substitution.} The model provides textbook-level mechanisms generally associated with the molecule (e.g., melatonin's antioxidant properties or receptor binding) instead of the specific experimental findings (e.g., precise gene expression patterns) required by the question. While the generated content is biologically plausible in a broad sense, it fails to match the ground truth details. A typical case is as follows:
\begin{promptbox}[Question]
How does melatonin exert anti-inflammatory effects on adult astrocytes?
\end{promptbox}

\begin{promptbox}[Correct Answer]
1. Melatonin reduces the expression of pro-inflammatory factors including NF-kappaB, COX-2, and iNOS.
2. It increases levels of cytokines IL-6 and IL-10, which are associated with anti-inflammatory responses.
3. Melatonin upregulates gene expression of protective transcription factors Nrf2 and SIRT1, indicating activation of anti-oxidative and anti-inflammatory pathways.
4. It modulates signaling by downregulating gene expression of AMPK and PGC-1alpha, although PGC-1alpha protein levels remain unchanged.
\end{promptbox}

\begin{promptbox}[Model Response (Generic Mechanism Substitution)]
1. Melatonin binds to astrocytic MT1/MT2 receptors, inhibiting the NF-kappaB pathway and subsequent pro-inflammatory cytokine transcription.
2. Its potent antioxidant action reduces intracellular reactive oxygen species, preventing the activation of the NLRP3 inflammasome.
3. Melatonin upregulates SIRT1 activity, which deacetylates the p65 subunit of NF-kappaB, thus silencing its transcriptional activity.
4. It enhances the expression and function of the astrocytic glutamate transporter GLT-1, thereby mitigating excitotoxicity-driven inflammatory signaling.
5. Melatonin reprograms astrocyte metabolism from a pro-inflammatory glycolytic state back towards mitochondrial oxidative phosphorylation.
\end{promptbox}

\textbf{Refusal to Answer.} Instead of attempting to infer a mechanism or hallucinate an answer, the model explicitly states that it cannot answer the question. This indicates a failure to retrieve relevant knowledge from its training data or a safety trigger that prevents engagement with the specific query. A typical case is as follows:

\begin{promptbox}[Model Response (Refusal to Answer)]
Sorry, I haven't learned how to answer this question yet. You can ask me some other questions and I'll do my best to help you solve them.
\end{promptbox}







\textbf{Overconfident Reasoning.} The model bypasses the tool-use step available in a ReAct framework and relies on its internal knowledge to infer a mechanism. While the reasoning process appears logical, it leads to a hallucinated answer because the ground truth is not part of the model's training data. A typical case is as follows:


\begin{promptbox}[Question]
What is the mechanism by which the glycoprotein SYB-1 from Syngnathoides biaculeatus enhances nonspecific immunity?
\end{promptbox}

\begin{promptbox}[Correct Answer]
1. SYB-1 significantly increases macrophage cell viability and promotes the release of key immune mediators including nitric oxide, TNF-alpha, and IL-6.
2. Metabolomic analysis indicates that its immunomodulatory activity is associated with the modulation of arachidonic acid metabolism.
3. Key molecular targets involved in this regulatory process are CYP450 enzyme family members and PTGS2.
4. This suggests that SYB-1 has potential as a natural functional food supplement for enhancing immune function.
\end{promptbox}

\begin{promptbox}[Model Response (Overconfident Reasoning)]
1. SYB-1 acts as a lectin-like opsonin that binds microbial surface glycans to agglutinate pathogens and mark them for hemocyte phagocytosis.
2. SYB-1 engagement of microbial carbohydrates initiates a lectin-like complement activation cascade that enhances pathogen lysis and uptake.
3. SYB-1 binds receptors on hemocytes (e.g., C-type lectin-type or scavenger receptors) to trigger signaling that upregulates proinflammatory cytokines and antimicrobial peptides.
4. SYB-1 directly stimulates phagocyte functions, increasing chemotaxis, phagocytic uptake, respiratory burst, and nitric oxide production for improved killing.
5. SYB-1 concentrates at mucosal surfaces to bridge microbes and immune cells, promoting local clearance and a sustained, trained-like elevation of innate responsiveness.
\end{promptbox}

\begin{promptbox}[Thinking Process (Overconfident Reasoning)]
SYB-1 is described as a glycoprotein from a teleost (Syngnathoides biaculeatus), so it likely functions in innate (nonspecific) immunity similar to known fish lectins and glycoproteins; teleost innate effectors act via pathogen binding (lectin/CRD activity), opsonization/agglutination, complement activation (including lectin pathway analogues), and by modulating hemocyte/macrophage responses (chemotaxis, phagocytosis, respiratory burst, and cytokine/AMP expression). Glycoproteins often present carbohydrate-recognition domains that bind PAMPs (LPS, beta-glucan, mannans), causing agglutination and enhanced phagocyte recognition. Binding to immune cell receptors (C-type lectin receptors, scavenger receptors, or TLR co-receptors) can trigger intracellular signaling to upregulate proinflammatory cytokines and antimicrobial peptides. Opsonization and lectin-mediated complement activation accelerate clearance by hemocytes and complement-mediated lysis, while direct stimulation of phagocyte oxidative/nitrosative responses raises microbicidal activity. SYB-1 might also act as a chemoattractant or adhesion bridge at mucosal surfaces to recruit and retain hemocytes and stabilize microbe-immune cell interactions, and could induce a trained-like heightened innate responsiveness after exposure.
\end{promptbox}

\section{Discussion and Conclusion}

\subsection{Discussion}

In this section, we discuss the key implications and future directions of our study.

\textbf{Performance Gap in Unknown Knowledge.}
Our evaluation results highlight a critical limitation of current LLMs: while they achieve impressive results on known knowledge (high performance on MMLU-Pro), their performance degrades significantly when confronted with unknown knowledge (low performance on our benchmark). The models excel at retrieving and synthesizing established facts but struggle to derive new knowledge that post-dates their training cutoff. This distinct performance gap indicates that LLMs still possess substantial room for improvement in new knowledge discovery tasks. It underscores that current models rely heavily on memorization and pattern matching rather than possessing the capability to truly reason about and assimilate novel scientific concepts from scratch.

\textbf{Generalization Ability of the Benchmark Pipeline.}
Addressing the challenges above requires robust evaluation tools. The methodological pipeline proposed in this study demonstrates strong generalization ability and is not confined to the biology domain. By simply replacing the source category in the JCR system, our systematic approach, spanning data acquisition, QA extraction, and QA filter, can be easily adapted to other scientific domains. This flexibility allows for the creation of a family of benchmarks, enabling a comprehensive, cross-disciplinary evaluation of how well AI models can master and discover new knowledge in various areas of human expertise.

Building on this study, we identify several promising avenues for future research: 
(1) The scope of model evaluation deserves expansion. Due to limits on time and money, we only select a representative set of models instead of testing all available SOTA LLMs. Future studies should evaluate a broader range of models to ensure a more comprehensive analysis. 
(2) Future research should develop a process-level evaluation framework. As an early attempt to assess the ability of LLMs to discover new knowledge, our current evaluation is primarily based on final outcomes. However, a model that arrives at a correct conclusion through flawed reasoning or logically invalid paths would still achieve a high score. Therefore, investigating the specific processes involved in knowledge discovery and proposing evaluation metrics based on these intermediate steps is essential for future research. 
(3) There is a need to develop novel methodologies to enhance LLMs' knowledge discovery ability. Our results indicate that existing LLMs still perform poorly on knowledge discovery tasks. To address this limitation, future work should explore and implement new methods, such as specialized training strategies or architectural innovations, to improve the knowledge discovery capabilities of LLMs.

\subsection{Conclusion}

In this paper, we propose DBench-Bio, a novel benchmark designed to rigorously evaluate the new knowledge discovery capabilities of LLMs within the biology domain. By establishing a systematic pipeline that extracts high-quality QA pairs from the latest academic literature and conducts multi-dimensional quality control, we provided a dynamic, contamination-free evaluation benchmark that reliably assesses complex scientific reasoning over new knowledge.
Our comprehensive evaluation reveals critical insights into the current limitations of LLMs. While existing models demonstrate strong proficiency in retrieving and reasoning over established knowledge, they face significant challenges when confronted with novel, unknown scientific findings that post-date their training cutoffs. 
A key strength of our work lies in the generalization ability of its methodological framework. The pipeline we proposed is inherently domain-agnostic; by simply replacing the source category within the JCR system, it can be directly adapted to construct benchmarks for other scientific domains, such as physics, chemistry, or social sciences. 
This flexibility allows for the creation of a family of benchmarks, enabling the community to track and compare AI-driven knowledge discovery progress across a broad spectrum of human expertise.
Looking forward, we hope DBench-Bio will serve as a foundational infrastructure to drive future research toward AI-driven knowledge discovery.

\begin{acks}
This research is supported by the NUS Artificial Intelligence Institute (NAII) Seed Grant (WBS: A-8003158-00-00), China Postdoctoral Science Foundation (No. BX20240186 and No. 2024M761616), and the Shuimu Tsinghua Scholar Program.
\end{acks}

\newpage
\bibliographystyle{ACM-Reference-Format}
\bibliography{main}

@String{Computer = "{IEEE} Computer" }

@article{jumper2021highly,
  title={Highly Accurate Protein Structure Prediction with AlphaFold},
  author={Jumper, John and Evans, Richard and Pritzel, Alexander and Green, Tim and Figurnov, Michael and Ronneberger, Olaf and Tunyasuvunakool, Kathryn and Bates, Russ and {\v{Z}}{\'\i}dek, Augustin and Potapenko, Anna and others},
  journal={Nature},
  volume={596},
  number={7873},
  pages={583--589},
  year={2021},
  publisher={Nature Publishing Group UK London}
}

@article{gao2024empowering,
  title={Empowering Biomedical Discovery with AI Agents},
  author={Gao, Shanghua and Fang, Ada and Huang, Yepeng and Giunchiglia, Valentina and Noori, Ayush and Schwarz, Jonathan Richard and Ektefaie, Yasha and Kondic, Jovana and Zitnik, Marinka},
  journal={Cell},
  volume={187},
  number={22},
  pages={6125--6151},
  year={2024},
  publisher={Elsevier}
}

@article{gottweis2025towards,
  title={Towards an AI Co-scientist},
  author={Gottweis, Juraj and Weng, Wei-Hung and Daryin, Alexander and Tu, Tao and Palepu, Anil and Sirkovic, Petar and Myaskovsky, Artiom and Weissenberger, Felix and Rong, Keran and Tanno, Ryutaro and others},
  journal={arXiv preprint arXiv:2502.18864},
  year={2025}
}

@article{huang2025biomni,
  title={Biomni: A General-purpose Biomedical AI Agent},
  author={Huang, Kexin and Zhang, Serena and Wang, Hanchen and Qu, Yuanhao and Lu, Yingzhou and Roohani, Yusuf and Li, Ryan and Qiu, Lin and Li, Gavin and Zhang, Junze and others},
  journal={biorxiv},
  year={2025}
}

@article{jin2025biolab,
  title={BioLab: End-to-End Autonomous Life Sciences Research with Multi-agents System Integrating Biological Foundation Models},
  author={Jin, Ruofan and Guo, Yucheng and Qu, Yuanhao and Yang, Ming and Shang, Chun and Yang, Qirong and Chao, Linlin and Zhou, Yi and Xu, Ruilai and Xu, Ziyao and others},
  journal={BioRxiv},
  pages={2025--09},
  year={2025},
  publisher={Cold Spring Harbor Laboratory}
}

@online{openai2025introducing,
  title        = {Introducing GPT-5.2},
  author       = {OpenAI},
  year         = 2025,
  url          = {https://openai.com/index/introducing-gpt-5-2/},
  urldate      = {2026-2-1}
}

@article{chen2025auto,
  title={Auto-Bench: An Automated Benchmark for Scientific Discovery in LLMs},
  author={Chen, Tingting and Anumasa, Srinivas and Lin, Beibei and Shah, Vedant and Goyal, Anirudh and Liu, Dianbo},
  journal={arXiv preprint arXiv:2502.15224},
  year={2025}
}

@article{liu2025researchbench,
  title={ResearchBench: Benchmarking LLMs in Scientific Discovery via Inspiration-Based Task Decomposition},
  author={Liu, Yujie and Yang, Zonglin and Xie, Tong and Ni, Jinjie and Gao, Ben and Li, Yuqiang and Tang, Shixiang and Ouyang, Wanli and Cambria, Erik and Zhou, Dongzhan},
  journal={arXiv preprint arXiv:2503.21248},
  year={2025}
}

@inproceedings{shojaee2025llmsrbench,
    title={LLM-SRBench: A New Benchmark for Scientific Equation Discovery with Large Language Models},
    author={Parshin Shojaee and Ngoc-Hieu Nguyen and Kazem Meidani and Amir Barati Farimani and Khoa D Doan and Chandan K. Reddy},
    booktitle={Forty-second International Conference on Machine Learning},
    year={2025},
    url={https://openreview.net/forum?id=SyQPiZJVWY}
}

@article{zhang2025hiscibench,
  title={HiSciBench: A Hierarchical Multi-disciplinary Benchmark for Scientific Intelligence from Reading to Discovery},
  author={Zhang, Yaping and Zhang, Qixuan and Zhang, Xingquan and Chen, Zhiyuan and Zhuang, Wenwen and Liang, Yupu and Xiang, Lu and Zhao, Yang and Zhang, Jiajun and Zhou, Yu and others},
  journal={arXiv preprint arXiv:2512.22899},
  year={2025}
}

@article{wang2024mmlu,
  title={MMLU-Pro: A More Robust and Challenging Multi-task Language Understanding Benchmark},
  author={Wang, Yubo and Ma, Xueguang and Zhang, Ge and Ni, Yuansheng and Chandra, Abhranil and Guo, Shiguang and Ren, Weiming and Arulraj, Aaran and He, Xuan and Jiang, Ziyan and others},
  journal={Advances in Neural Information Processing Systems},
  volume={37},
  pages={95266--95290},
  year={2024}
}

@article{wang2019superglue,
  title={SuperGLUE: A Stickier Benchmark for General-Purpose Language Understanding Systems},
  author={Wang, Alex and Pruksachatkun, Yada and Nangia, Nikita and Singh, Amanpreet and Michael, Julian and Hill, Felix and Levy, Omer and Bowman, Samuel},
  journal={Advances in neural information processing systems},
  volume={32},
  year={2019}
}

@article{phan2025humanity,
  title={Humanity's Last Exam},
  author={Phan, Long and Gatti, Alice and Han, Ziwen and Li, Nathaniel and Hu, Josephina and Zhang, Hugh and Zhang, Chen Bo Calvin and Shaaban, Mohamed and Ling, John and Shi, Sean and others},
  journal={arXiv preprint arXiv:2501.14249},
  year={2025}
}

@inproceedings{rein2024gpqa,
  title={GPQA: A Graduate-Level Google-Proof Q\&A Benchmark},
  author={Rein, David and Hou, Betty Li and Stickland, Asa Cooper and Petty, Jackson and Pang, Richard Yuanzhe and Dirani, Julien and Michael, Julian and Bowman, Samuel R},
  booktitle={First Conference on Language Modeling},
  year={2024}
}

@inproceedings{long2024llms,
  title={On LLMs-Driven Synthetic Data Generation, Curation, and Evaluation: A Survey},
  author={Long, Lin and Wang, Rui and Xiao, Ruixuan and Zhao, Junbo and Ding, Xiao and Chen, Gang and Wang, Haobo},
  booktitle={ACL (Findings)},
  year={2024}
}

@inproceedings{chen2025benchmarking,
  title={Benchmarking Large Language Models Under Data Contamination: A Survey from Static to Dynamic Evaluation},
  author={Chen, Simin and Chen, Yiming and Li, Zexin and Jiang, Yifan and Wan, Zhongwei and He, Yixin and Ran, Dezhi and Gu, Tianle and Li, Haizhou and Xie, Tao and others},
  booktitle={Proceedings of the 2025 Conference on Empirical Methods in Natural Language Processing},
  pages={10091--10109},
  year={2025}
}

@inproceedings{white2025livebench,
  title={LiveBench: A Challenging, Contamination-Limited LLM Benchmark},
  author={White, Colin and Dooley, Samuel and Roberts, Manley and Pal, Arka and Feuer, Benjamin and Jain, Siddhartha and Shwartz-Ziv, Ravid and Jain, Neel and Saifullah, Khalid and Dey, Sreemanti and others},
  booktitle={The Thirteenth International Conference on Learning Representations},
  year={2025}
}

@inproceedings{jain2025livecodebench,
  title={LiveCodeBench: Holistic and Contamination Free Evaluation of Large Language Models for Code},
  author={Jain, Naman and Han, King and Gu, Alex and Li, Wen-Ding and Yan, Fanjia and Zhang, Tianjun and Wang, Sida and Solar-Lezama, Armando and Sen, Koushik and Stoica, Ion},
  booktitle={The Thirteenth International Conference on Learning Representations},
  year={2025}
}

@inproceedings{karger2025forecastbench,
  title={ForecastBench: A Dynamic Benchmark of AI Forecasting Capabilities},
  author={Karger, Ezra and Bastani, Houtan and Yueh-Han, Chen and Jacobs, Zachary and Halawi, Danny and Zhang, Fred and Tetlock, Philip},
  booktitle={The Thirteenth International Conference on Learning Representations},
  year={2025}
}

@article{zeng2025futurex,
  title={FutureX: An Advanced Live Benchmark for LLM Agents in Future Prediction},
  author={Zeng, Zhiyuan and Liu, Jiashuo and Chen, Siyuan and He, Tianci and Liao, Yali and Tian, Yixiao and Wang, Jinpeng and Wang, Zaiyuan and Yang, Yang and Yin, Lingyue and others},
  journal={arXiv preprint arXiv:2508.11987},
  year={2025}
}

@inproceedings{hendrycks2021mmlu,
    title={Measuring Massive Multitask Language Understanding},
    author={Dan Hendrycks and Collin Burns and Steven Basart and Andy Zou and Mantas Mazeika and Dawn Song and Jacob Steinhardt},
    booktitle={International Conference on Learning Representations},
    year={2021},
    url={https://openreview.net/forum?id=d7KBjmI3GmQ}
}

@inproceedings{yue2024mmmu,
  title={MMMU: A Massive Multi-discipline Multimodal Understanding and Reasoning Benchmark for Expert AGI},
  author={Yue, Xiang and Ni, Yuansheng and Zhang, Kai and Zheng, Tianyu and Liu, Ruoqi and Zhang, Ge and Stevens, Samuel and Jiang, Dongfu and Ren, Weiming and Sun, Yuxuan and others},
  booktitle={Proceedings of the IEEE/CVF Conference on Computer Vision and Pattern Recognition},
  pages={9556--9567},
  year={2024}
}

@article{huang2023c,
  title={C-Eval: A Multi-level Multi-discipline Chinese Evaluation Suite for Foundation Models},
  author={Huang, Yuzhen and Bai, Yuzhuo and Zhu, Zhihao and Zhang, Junlei and Zhang, Jinghan and Su, Tangjun and Liu, Junteng and Lv, Chuancheng and Zhang, Yikai and Fu, Yao and others},
  journal={Advances in Neural Information Processing Systems},
  volume={36},
  pages={62991--63010},
  year={2023}
}

@inproceedings{gu2024xiezhi,
  title={Xiezhi: An Ever-Updating Benchmark for Holistic Domain Knowledge Evaluation},
  author={Gu, Zhouhong and Zhu, Xiaoxuan and Ye, Haoning and Zhang, Lin and Wang, Jianchen and Zhu, Yixin and Jiang, Sihang and Xiong, Zhuozhi and Li, Zihan and Wu, Weijie and others},
  booktitle={Proceedings of the AAAI conference on artificial intelligence},
  volume={38},
  number={16},
  pages={18099--18107},
  year={2024}
}

@inproceedings{sun2024scieval,
  title={SciEval: A Multi-level Large Language Model Evaluation Benchmark for Scientific Research},
  author={Sun, Liangtai and Han, Yang and Zhao, Zihan and Ma, Da and Shen, Zhennan and Chen, Baocai and Chen, Lu and Yu, Kai},
  booktitle={Proceedings of the AAAI Conference on Artificial Intelligence},
  volume={38},
  number={17},
  pages={19053--19061},
  year={2024}
}

@article{huang2024olympicarena,
  title={OlympicArena: Benchmarking Multi-discipline Cognitive Reasoning for Superintelligent AI},
  author={Huang, Zhen and Wang, Zengzhi and Xia, Shijie and Li, Xuefeng and Zou, Haoyang and Xu, Ruijie and Fan, Run-Ze and Ye, Lyumanshan and Chern, Ethan and Ye, Yixin and others},
  journal={Advances in Neural Information Processing Systems},
  volume={37},
  pages={19209--19253},
  year={2024}
}

@article{jia2025towards,
  title={Towards Adaptive ML Benchmarks: Web-Agent-Driven Construction, Domain Expansion, And Metric Optimization},
  author={Jia, Hangyi and Qian, Yuxi and Tong, Hanwen and Wu, Xinhui and Chen, Lin and Wei, Feng},
  journal={arXiv preprint arXiv:2509.09321},
  year={2025}
}

@inproceedings{mirzadeh2024gsm,
  title={GSM-Symbolic: Understanding the Limitations of Mathematical Reasoning in Large Language Models},
  author={Mirzadeh, Seyed Iman and Alizadeh, Keivan and Shahrokhi, Hooman and Tuzel, Oncel and Bengio, Samy and Farajtabar, Mehrdad},
  booktitle={The Thirteenth International Conference on Learning Representations},
  year={2024}
}

@inproceedings{kurtic2024mathador,
  title={Mathador-LM: A Dynamic Benchmark for Mathematical Reasoning on Large Language Models},
  author={Kurtic, Eldar and Moeini, Amir and Alistarh, Dan},
  booktitle={Proceedings of the 2024 Conference on Empirical Methods in Natural Language Processing},
  pages={17020--17027},
  year={2024}
}

@article{wan2024sciqag,
  title={SciQAG: A Framework for Auto-Generated Science Question Answering Dataset with Fine-Grained Evaluation},
  author={Wan, Yuwei and Liu, Yixuan and Ajith, Aswathy and Grazian, Clara and Hoex, Bram and Zhang, Wenjie and Kit, Chunyu and Xie, Tong and Foster, Ian},
  journal={arXiv preprint arXiv:2405.09939},
  year={2024}
}

@inproceedings{yao2022react,
  title={ReAct: Synergizing Reasoning and Acting in Language Models},
  author={Yao, Shunyu and Zhao, Jeffrey and Yu, Dian and Du, Nan and Shafran, Izhak and Narasimhan, Karthik R and Cao, Yuan},
  booktitle={The eleventh international conference on learning representations},
  year={2022}
}

@inproceedings{calderon-etal-2025-alternative,
    title = "The Alternative Annotator Test for LLM-as-a-Judge: How to Statistically Justify Replacing Human Annotators with LLMs",
    author = "Calderon, Nitay  and
      Reichart, Roi  and
      Dror, Rotem",
    booktitle = "Proceedings of the 63rd Annual Meeting of the Association for Computational Linguistics (Volume 1: Long Papers)",
    month = jul,
    year = "2025",
    address = "Vienna, Austria",
    publisher = "Association for Computational Linguistics",
    url = "https://aclanthology.org/2025.acl-long.782/",
    doi = "10.18653/v1/2025.acl-long.782",
    pages = "16051--16081",
    ISBN = "979-8-89176-251-0",
}

@article{gu2024survey,
  title={A Survey on LLM-as-a-judge},
  author={Gu, Jiawei and Jiang, Xuhui and Shi, Zhichao and Tan, Hexiang and Zhai, Xuehao and Xu, Chengjin and Li, Wei and Shen, Yinghan and Ma, Shengjie and Liu, Honghao and others},
  journal={The Innovation},
  publisher={Elsevier},
  year={2024}
}

@article{li2024llms,
  title={LLMs-as-judges: A Comprehensive Survey on LLM-Based Evaluation Methods},
  author={Li, Haitao and Dong, Qian and Chen, Junjie and Su, Huixue and Zhou, Yujia and Ai, Qingyao and Ye, Ziyi and Liu, Yiqun},
  journal={arXiv preprint arXiv:2412.05579},
  year={2024}
}

@article{arslan2024survey,
  title={A Survey on RAG with LLMs},
  author={Arslan, Muhammad and Ghanem, Hussam and Munawar, Saba and Cruz, Christophe},
  journal={Procedia Computer Science},
  volume={246},
  pages={3781--3790},
  year={2024},
  publisher={Elsevier}
}

@book{hanson1979patterns,
  title={Patterns of discovery: An inquiry into the conceptual foundations of science},
  author={Hanson, Norwood Russell},
  year={1979},
  publisher={Cup archive}
}

@book{dunbar1995scientists,
  title={How scientists really reason: Scientific reasoning in real-world laboratories.},
  author={Dunbar, Kevin},
  year={1995},
  publisher={The MIT Press}
}

\appendix
\section{Appendix}

\subsection{List of 12 Sub-domains}
\begin{enumerate}[leftmargin=1.5em,label=\arabic*.]
    \item Biochemistry \& Molecular Biology
    \item Biophysics
    \item Biotechnology \& Applied Microbiology
    \item Cell \& Tissue Engineering
    \item Cell Biology
    \item Chemistry, Medicinal
    \item Genetics \& Heredity
    \item Mathematical \& Computational Biology
    \item Neurosciences
    \item Pathology
    \item Pharmacology \& Pharmacy
    \item Physiology
\end{enumerate}

\subsection{Model Details}
The details of the model we use are as follows:
\begin{itemize}
    \item GPT-5: Released on August 7, 2025.
    \item GPT-5-Mini: Released on August 7, 2025.
    \item GPT-5.2: Released on December 11, 2025.
    \item Gemini-3-Pro: Released on November 18, 2025. Does not support disabling the thinking mode.
    \item Gemini-3-Flash: Released on December 18, 2025.
    \item DeepSeek-V3.2: Released on December 1, 2025.
    \item Kimi-K2: The non-thinking model was released on July 11, 2025; the thinking model was released on November 6, 2025.
    \item GLM-4.6: Released on September 30, 2025.
    \item GLM-4.7: Released on December 23, 2025.
    \item Qwen3-Max: The official non-thinking model was released on September 24, 2025; the official thinking model was released on January 26, 2026.
    \item Baichuan-M2: Released on August 11, 2025.
\end{itemize}

\subsection{Prompts}
\label{app:prompts}

\begin{promptbox}[Prompt for Generating QA Pairs]
# Task
Based on the provided scientific paper abstract, transform its main scientific discovery into a single, high-quality short-answer question-and-answer (QA) pair that strictly adhere to the Guidelines below.

# Guidelines
1. The question corresponds to the main scientific question of the abstract, and the answer corresponds to the scientific discovery result.
2. **Focus on Scientific Discovery**: Examples:
    - **Causal Inference**: e.g., "Does X regulate Y phenomenon?" or "What are the implications of Z on disease progression?"
    - **Mechanistic Explanation**: e.g., "How does X regulate Y phenomenon?" or "What is the mechanism behind Z effect?"
3. **Clarity and Natural Phrasing**: 
    - Questions and answers must be self-contained, written in natural English. Avoid phrases like "based on the text" or "in this study". 
    - Answers should be bullet-style, clear and concise.
    - Focus on high-level concepts, and avoid fine-grained details like specific parameters or numerical values. 

# Input
paper abstract

# Output JSON Schema
{
    "question": "full English question ending with ?",
    "answer": "1. first bullet point.\\n2. second bullet point.\\n...",
}

# Example

## Input
Cancer cells with RAS mutations exhibit enhanced autophagy, essential for their proliferation and survival, making it a potential target for therapeutic intervention. However, the regulatory differences between RAS-induced autophagy and physiological autophagy remain poorly understood, complicating the development of cancer-specific anti-autophagy treatments. In this study, we identified a form of non-canonical autophagy induced by oncogenic KRAS expression, termed RAS-induced non-canonical autophagy via ATG8ylation (RINCAA). RINCAA involves distinct autophagic factors compared to those in starvation-induced autophagy and incorporates non-autophagic components, resulting in the formation of non-canonical autophagosomes with multivesicular/multilaminar structures labeled by ATG8 family proteins (e.g., LC3 and GABARAP). We have designated these structures as RAS-induced multivesicular/multilaminar bodies of ATG8ylation (RIMMBA). A notable feature of RINCAA is the substitution of the class III PI3K in canonical autophagy with PI4KB in RINCAA. We identified a regulatory P38-ULK1-PI4KB-WIPI2 signaling cascade governing this process, where ULK1 triggers PI4KB phosphorylation at S256 and T263, initiating PI4P production, ATG8ylation, and non-canonical autophagy. Importantly, elevated PI4KB phosphorylation at S256 and T263 was observed in RASmutated cancer cells and colorectal cancer specimens. Inhibition of PI4KB S256 and T263 phosphorylation led to a reduction in RINCAA activity and tumor growth in both xenograft and KPC models of pancreatic cancer, suggesting that targeting ULK1mediated PI4KB phosphorylation could represent a promising therapeutic strategy for RAS-mutated cancers.

## Output
{
    "question": "Is there any mechanistic difference between RAS-induced autophagy and physiological autophagy?",
    "answer": "1. Oncogenic RAS induces a specific form of non-canonical autophagy, which is distinct from physiological autophagy.\\n2. A key molecular difference is the substitution of the class III PI3K, essential for canonical autophagy, with the enzyme PI4KB in the RAS-induced pathway.\\n3. The process is governed by a unique signaling cascade where P38 activates ULK1, which in turn phosphorylates PI4KB at specific sites (S256 and T263) to initiate the autophagic process.\\n4. It produces atypical autophagosomes with multivesicular/multilaminar structures labeled by ATG8 family proteins.",
}
\end{promptbox}

\begin{promptbox}[Prompt for Relevance]
# ROLE: Field Relevance Scorer
You are an expert specializing in {field}.  
Your task is to evaluate the relevance of each (question, answer) pair to the field of {field}.

## Scoring scale (1-5):
5 = Highly central: The topic is primarily studied within {field}.  
4 = Directly related: The content is clearly within the scope of {field}.
3 = Moderately related: {field} is one of several equally important contexts, or it is clearly implicated but not the main focus.  
2 = Minimally related: {field} is mentioned only tangentially or as a minor downstream effect.  
1 = Not related: The content has no meaningful connection to {field}.

# INPUT FORMAT:
| id | question | answer |
|---|---|---|
| <id> | <question> | <answer> |

# OUTPUT FORMAT:
| id | score |
|---|---|
| <id> | <score 1-5> |

# NOTE:
- Provide only the output table. Do not include explanations or additional columns.
\end{promptbox}

\begin{promptbox}[Prompt for Clarity]
# TASK
Access the clarity of each (question, answer) pair. The question and answer should be self-contained, in natural English, and free of ambiguity. No strange words, such as "in the text", "according to the figure", or "based on the table".

## Scoring scale (1-5):
5 = Excellent. Fully natural, clear, and unambiguous.
4 = Good. Natural and clear, with only minor stiffness in phrasing. 
3 = Fair. Generally understandable but contains unnatural phrasing or slight ambiguity. 
2 = Poor. Significant clarity issues.
1 = Unacceptable. Confusing, incomprehensible, severely ambiguous.

# INPUT FORMAT:
| id | question | answer |
|---|---|---|
| <id> | <question> | <answer> |

# OUTPUT FORMAT:
| id | score |
|---|---|
| <id> | <score 1-5> |

# NOTE:
- Provide only the output table. Do not include explanations or additional columns.
\end{promptbox}

\begin{promptbox}[Prompt for Centrality]
# TASK
Your task is to assess the centrality of each (question, answer) pair. Centrality measures whether the QA pair focuses on the main scientific discovery of the abstract (e.g., the primary finding, proposed mechanism, central causal relationship, or key conclusion), rather than on secondary or supporting details.

## Scoring scale (1-5):
5 = Excellent. Directly addresses the core discovery.
4 = Good. Focuses on a major aspect or direct consequence of the core discovery.
3 = Fair. Peripheral or partially related to the core discovery.
2 = Poor. Focuses on secondary or minor details.
1 = Unacceptable. Not related to the core discovery.

# INPUT FORMAT:
| id | abstract | question | answer |
|---|---|---|---|
| <id> | <abstract> | <question> | <answer> |

# OUTPUT FORMAT:
| id | score |
|---|---|
| <id> | <score 1-5> |

# NOTE:
- Provide only the output table. Do not include explanations or additional columns.
\end{promptbox}

\begin{promptbox}[Prompt for Evaluation]
s# Role
You are a senior scientist with deep expertise.

# Task
The user will input a reference answer and a candidate answer, you should evaluate the quality of the candidate answer based on the reference answer, and output a JSON object containing a reasoned explanation and a score.

# Scoring Criteria
- 5 (Perfect): The answer is entirely correct, logically rigorous, and comprehensive. It is highly consistent with the reference answer on all core facts.
- 4 (Excellent): The answer is correct on core facts and logically sound, but may have minor flaws in detail, completeness, or phrasing.
- 3 (Passable): The answer contains some correct information but also has significant factual errors, logical inconsistencies, or is missing key information.
- 2 (Poor): The answer is largely incorrect or irrelevant, failing to accurately address the question.
- 1 (Incorrect): The answer is completely wrong or contains fundamental scientific fallacies.

# Output JSON Format
{
    "reason": str,  // A detailed explanation of the score, highlighting strengths and weaknesses of the candidate answer.
    "score": int  // An integer score from 1 to 5 based on the criteria above.
}

NOTE: Return ONLY a valid JSON object, WITHOUT markdown code blocks or backticks.
\end{promptbox}

\end{document}